%% file: main.tex
\documentclass[letterpaper, 10 pt, journal, twoside]{IEEEtran}
\input{preamble}

\begin{document}

\title{Unified Complementarity-Based Contact Modeling and Planning for Soft Robots}

\author{Milad Azizkhani and Yue Chen%

\thanks{Project page: \url{https://miladroboticist.github.io/soft-robot-contact-planning/}. 
M. Azizkhani is with the Woodruff School of Mechanical Engineering and the Institute for Robotics and Intelligent Machines (IRIM), Georgia Institute of Technology, Atlanta, GA, USA (e-mail: mazizkhani3@gatech.edu). 
Y. Chen is with the Coulter Department of Biomedical Engineering (Georgia Tech and Emory University) and the Institute for Robotics and Intelligent Machines (IRIM), Georgia Institute of Technology, Atlanta, GA, USA (e-mail: yue.chen@bme.gatech.edu). 
This work was supported in part by the National Science Foundation (NSF) CAREER Award~2339202 and in part by the USDA-NIFA National Robotics Initiative through the National Science Foundation under Award~2022-11065.}}

\maketitle
\vspace{-5 mm}

\begin{abstract}

  Soft robots were introduced in large part to enable safe, adaptive interaction with the environment, and this interaction relies fundamentally on contact. However, modeling and planning contact-rich interactions for soft robots remain challenging: dense contact candidates along the body create redundant constraints and rank-deficient LCPs, while the disparity between high stiffness and low friction introduces severe ill-conditioning. Existing approaches rely on problem-specific approximations or penalty-based treatments. This letter presents a unified complementarity-based framework for soft-robot contact modeling and planning that brings contact modeling, manipulation, and planning into a unified, physically consistent formulation. We develop a robust Linear Complementarity Problem (LCP) model tailored to discretized soft robots and address these challenges with a three-stage conditioning pipeline: inertial rank selection to remove redundant contacts, Ruiz equilibration to correct scale disparity and ill-conditioning, and lightweight Tikhonov regularization on normal blocks. Building on the same formulation, we introduce a kinematically guided warm-start strategy that enables dynamic trajectory optimization through contact using Mathematical Programs with Complementarity Constraints (MPCC) and demonstrate its effectiveness on contact-rich ball manipulation tasks. In conclusion, CUSP provides a new foundation for unifying contact modeling, simulation, and planning in soft robotics.
\end{abstract}


\IEEEpeerreviewmaketitle

\section{Introduction}

Soft robots have emerged as a new class of robotic systems that combine continuum structures with intrinsic compliance, enabling them to navigate confined spaces, adapt to unstructured environments, and interact safely with humans and delicate objects \cite{della2023model, laschi2016soft, qiu2023tendon}. Nature illustrates that the full potential of compliant systems arises not only from intrinsic compliance but from whole-body, multi-contact interactions: elephants manipulate heavy logs with their trunks, starfish pry open shells by distributing force across multiple arms, and octopuses coordinate multiple tentacles in cluttered spaces \cite{leanza2024elephant,tekinalp2024topology}. Such contact-rich strategies enable tasks that would be impossible with conventional fingertip grasps or free-space motions. Translating these capabilities into soft robotics requires treating contact as a resource of ``passive actuation input'' rather than an obstacle to avoid, and thus two complementary capabilities: (i) a physics model and simulation environment that capture whole-body, multi-contact interactions in soft robots, and (ii) a mathematical formulation that enables planning through such contacts. While rigid-body systems benefit from well-established Linear Complementarity Problem (LCP) formulations for contact dynamics \cite{anitescu1997formulating,anitescu2006optimization,stewart1996implicit}, extending them to soft robots remains an open challenge. In this work, we address this gap by formulating an LCP-based framework tailored to soft robots and demonstrate its capability for both stable forward simulation and contact-rich trajectory optimization within a unified framework.

\vspace{-4 mm}
\subsection{Related Work}
\subsubsection{Soft Robot Modeling}

Soft robots can be modeled using high-fidelity approaches such as finite element \cite{duriez2013control, largilliere2015real} or Cosserat rod formulations \cite{renda2014dynamic, till2019real}. While these methods capture detailed deformations, they require fine discretization and repeated solution of large nonlinear systems, which makes them too slow for high-speed simulation, planning, or control. Strain-parameterized models offer some simplification compared to the standard Cosserat rod models, but they remain computationally demanding in practice \cite{boyer2020dynamics, renda2020geometric}. A more tractable alternative is the piecewise constant curvature (PCC) assumption \cite{godage2016dynamics, azizkhani2022dynamic}, where each section is approximated by simple geometric primitives that capture bending and elongation, yielding compact, low-dimensional dynamics that can be evaluated efficiently and integrated into optimization, with residual modeling errors addressed by adaptive or robust control \cite{azizkhani2023dynamic, azizkhani2025dynamic, kazemipour2022adaptive}.

\subsubsection{Contact Dynamics}

Contact modeling introduces nonlinearity, discontinuity, and nonsmooth transitions into robot dynamics. Penalty-based methods are straightforward to implement but require careful tuning and fail to represent sticking–sliding friction transitions consistently \cite{anitescu1997formulating,todorov2012mujoco}; complementarity-based formulations express non-penetration, unilateral contact forces, and Coulomb friction as complementary variables in a linear complementarity problem (LCP), ensuring exact constraint enforcement and structured friction treatment, and have become the state of the art for rigid-body simulation and trajectory optimization through contact \cite{stewart1996implicit,anitescu1997formulating,anitescu2006optimization,posa2014direct}.

\subsubsection{Soft Robot Contact Modeling}

In soft robotics, contact models have generally remained simplified: penalty-based approaches have been used, e.g., penalizing node penetrations in trajectory optimization for cable-driven soft robots \cite{bern2019trajectory}, and finite element method (FEM) frameworks have incorporated contact \cite{coevoet2017optimization,coevoet2019soft}, but the latter are typically used for high-fidelity deformation and inverse simulation rather than contact-rich manipulation and planning. FEM-based contact has been limited to proof-of-concept demonstrations on simple robots with few degrees of freedom (DoF) and to sticking without sliding friction; as DoF and geometric complexity increase, FEM becomes computationally expensive for trajectory optimization and even forward simulation, and is difficult to integrate with feedback control. Reduced-order models in ordinary differential equation form are more compatible with feedback design and essential for manipulation; capturing all relevant contact modes (sticking, sliding, and separation) requires scalable, reduced-order, complementarity-based formulations for soft robotic manipulators.

\vspace{-3 mm}
\subsection{Contributions of the Proposed Work}

This letter introduces (i) a \textbf{unified complementarity-based contact framework for soft robots}: a single contact model underlies both forward simulation and trajectory optimization, so that the same contact mechanics apply whether one is simulating forward in time or planning a trajectory through contact.
(ii)~\textbf{A robust LCP formulation for discretized soft robots.} By discretized soft robots we mean that contact is evaluated at predefined locations (disks) along the body rather than at every point on the continuum; the number of contact candidates can be chosen based on design (see Fig.~\ref{fig:soft_robot} and \cite{azizkhani2025dynamic} for the robot design). We identify a critical modeling challenge: representing a continuum body with dense contact disks creates redundant constraints that render the resulting LCP matrix singular or severely ill-conditioned. We resolve this by proposing a three-stage conditioning pipeline, inertial rank-revealing QR, Ruiz equilibration, and lightweight Tikhonov regularization, applied before the LCP solve. This conditioning yields a well-behaved matrix that can be solved reliably by standard LCP solvers, independent of the particular solver used. Unlike prior soft-robot formulations restricted to sticking or penalty-based approximations, our approach captures the full hybrid evolution of separation, sticking, and sliding friction in dense multi-contact settings.
(iii)~\textbf{Hierarchical optimization for trajectory planning.} To address the difficulty of solving large-scale mathematical programs with complementarity constraints (MPCC) for soft robots, we propose a hierarchical strategy. We first solve a kinodynamic planning problem to obtain a feasible contact strategy, then warm-start the full dynamic optimization with this solution. This guidance allows the full dynamic stage to converge to a more optimal, dynamically feasible trajectory for contact-rich manipulation tasks that are otherwise difficult to solve from a cold start.
To facilitate adoption by the soft robotics community, the proposed framework is designed for modularity, allowing users to define complex contact environments with minimal setup.
The complete implementation code will be made open-source upon acceptance.

\section{Background} 
\label{sec:Background}

\subsection{Soft-Robot Modeling}
\label{sec:modeling}

\begin{figure}[t]
  \centering
  \includegraphics[width=0.49\textwidth]{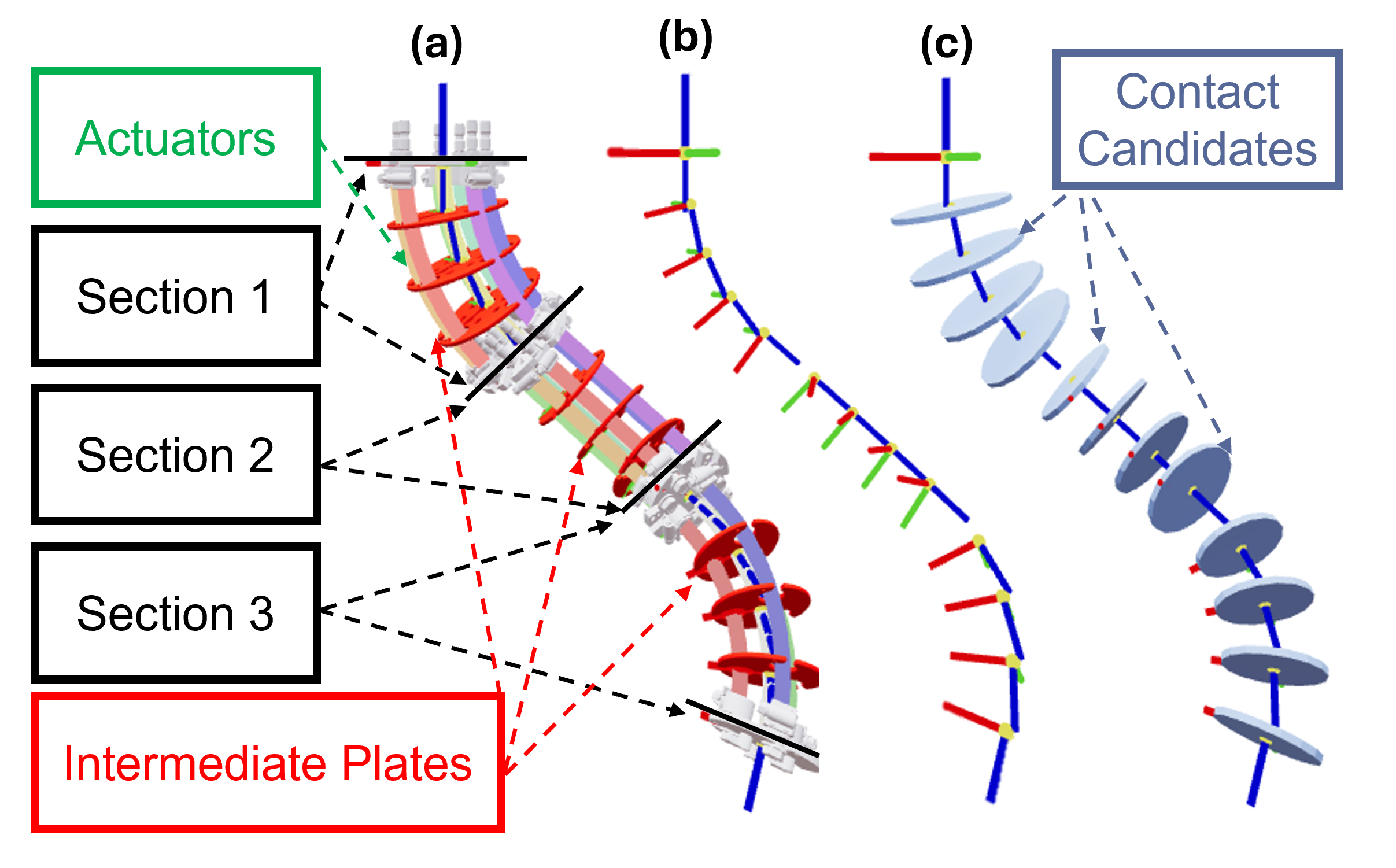}
  \caption{Three-section pneumatically actuated soft robot used in this work. 
  (a) Each section has three actuation units and intermediate plates (red) enforcing the piecewise constant curvature model. 
  (b) Local frames along the backbone represent the robot's deformation. 
  (c) Top, bottom, and intermediate plates are modeled as convex disks (light blue) serving as contact candidates.}
  \label{fig:soft_robot}
  \vspace{-5 mm}
\end{figure}

We consider a three-section pneumatically actuated soft manipulator modeled under the PCC assumption 
\cite{azizkhani2025dynamic,godage2016dynamics}, depicted in Fig.~\ref{fig:soft_robot}.  
Each section is driven by three chambers, giving nine actuator variables.
While we demonstrate the framework on this specific robot, the approach generalizes to any number of sections and contact candidates.
\begin{equation}
\boldsymbol{l} = [l_{1,1},l_{1,2},l_{1,3}, \; l_{2,1},l_{2,2},l_{2,3}, \; l_{3,1},l_{3,2},l_{3,3}]^\top \in \mathbb{R}^9,
\end{equation}
where $l_{i,j}$ denotes the elongation of chamber $j$ in section $i$.  
These actuator-space variables serve as the generalized coordinates for dynamics. Using these coordinates and the PCC formulation, we derive two mappings: from actuator space to configuration space and from configuration space to task space. By the chain rule, the position and orientation of any point along the robot body can then be obtained in closed form \cite{azizkhani2025soft}.
The equations of motion in actuator space are
\begin{equation}
\mathbf{M}(\boldsymbol{l})\,\ddot{\boldsymbol{l}} +
\boldsymbol{C}(\boldsymbol{l},\dot{\boldsymbol{l}})\,\dot{\boldsymbol{l}} +
\boldsymbol{K}\,\boldsymbol{l} +
\boldsymbol{B}\,\dot{\boldsymbol{l}} +
\boldsymbol{G}(\boldsymbol{l})
= \boldsymbol{J}^\top \mathbf{F}_{\text{ext}} + \boldsymbol{\tau},
\label{eq:EoM}
\end{equation}
where $\mathbf{M}$ is the inertia matrix, $\boldsymbol{C}$ the Coriolis/centrifugal terms, $\boldsymbol{K}$ stiffness, $\boldsymbol{B}$ damping, $\boldsymbol{G}$ gravity, $\mathbf{F}_{\text{ext}}$ external contact forces, $\boldsymbol{J}$ the contact Jacobian, and $\boldsymbol{\tau}$ the actuation input.  
We adopt standard approximations for tractability: rotational kinetic energy is small relative to translational, Coriolis/centrifugal effects are modest, and elastic behavior dominates \cite{falkenhahn2015dynamic,della2023model}.  
Full derivations and implementation details are provided in \cite{azizkhani2025dynamic}.
\vspace{-4 mm}
\subsection{LCP Formulation}
\label{sec:LCPbackground}

The LCP is defined as
\begin{equation}
\mathbf{w} = \mathbf{A}\mathbf{z} + \mathbf{b}, \quad \mathbf{w} \geq 0, \quad \mathbf{z} \geq 0, \quad \mathbf{w}^\top \mathbf{z} = 0,
\label{eq:LCP_def}
\end{equation}
where $\mathbf{A}\in\mathbb{R}^{n\times n}$, $\mathbf{b}\in\mathbb{R}^n$, and the solution is given by vectors $\mathbf{w},\mathbf{z}\in\mathbb{R}^n$.
% \yf{put constraints on the same line to save space}
The complementarity condition $\mathbf{w}^\top \mathbf{z}=0$ enforces mutually exclusive states: for each index $i$, either $w_i>0$ and $z_i=0$, or $z_i>0$ and $w_i=0$, or both are zero.  
This makes the LCP well-suited for representing systems with switching behaviors such as active/inactive constraints, forming the basis for contact formulations in this work.

\vspace{-3 mm}
\section{Forward Simulation for Robot Contact Modeling  via Robust LCP %\yc{Robot Contact Modeling via Robust LCP}
}
\label{sec:Forward}

Simulating contact for soft robots faces a fundamental numerical challenge: discretizing a continuum body into multiple contact candidates (e.g., disks) can introduce redundancy, in the sense that multiple discrete contact points impose  dependent constraints on the limited set of generalized coordinates. As a result, the resulting contact Jacobians may become rank-deficient, rendering standard LCP formulations structurally singular and leading to non-unique solutions or solver divergence unless properly conditioned.

This section presents a robust framework to resolve these pathologies. We integrate the dynamic model from Section~\ref{sec:modeling} with a specialized LCP conditioning pipeline. By incorporating inertial rank-selection prior to the solver call, we ensure numerical stability and physical consistency, enabling high-fidelity simulation of dense multi-contact interactions that are otherwise intractable for standard physics engines.
\vspace{-4 mm}
\subsection{Overview and Modeling Objective}

The objective of our simulation framework is to model contact interactions for soft robotic systems in a physically consistent manner.  
To this end, the contact formulation must satisfy two fundamental requirements:  
(i) \textit{nonpenetration}, ensuring that contact forces arise only when the local gap between the robot and its environment closes, and  
(ii) \textit{Coulomb friction}, constraining tangential forces within a friction cone defined by the normal force and friction coefficient.  
Together, these conditions not only prevent interpenetration but also determine the contact modes, namely sticking, sliding, or separation, within a unified complementarity formulation.

\vspace{-4 mm}
\subsection{Complementarity Formulation}
\label{subsec:CompForm}

In contact modeling, constraints are typically introduced to prevent interpenetration between two bodies. Let $g_i(\boldsymbol{l}) \ge 0$ denote the signed normal gap function for a contact candidate $i$, where $\boldsymbol{l}$ is the configuration vector of the system. When $g_i > 0$, the bodies are separated; when $g_i = 0$, contact occurs. The rate of change of this gap is $\dot{g}_i = \nabla_{\boldsymbol{l}} g_i(\boldsymbol{l})^\top \dot{\boldsymbol{l}}$, which gives the relative normal velocity $\nu_{n,i}$ at the contact point.

Traditionally, position-level constraints of the form $g_i(\boldsymbol{l}) \ge 0$ have been used to enforce nonpenetration.  
However, enforcing these constraints directly requires continuous contact detection and often introduces impulsive reactions  when transitions occur.  
To avoid such discontinuities, time-stepping formulations such as those of Stewart and Trinkle~\cite{stewart1996implicit} and Anitescu and Potra~\cite{anitescu1997formulating} impose complementarity on the \textit{contact velocity} instead of on the positional gap.  
This approach requires three complementarity conditions to fully characterize contact modes. First, a compressive normal force $f_{n,i}$ may only exist when the normal relative velocity is zero:
\begin{equation}
0 \le \nu_{n,i} \perp f_{n,i} \ge 0.
\label{eq:vel_comp_base}
\end{equation}
If the contact is separating ($\nu_{n,i} > 0$), no normal force acts ($f_{n,i} = 0$); if a compressive force is active ($f_{n,i} > 0$), the relative normal velocity must vanish.  
This velocity-level formulation allows stable, event-free integration and avoids impulsive corrections caused by explicit contact enforcement.

To describe tangential interactions, we adopt the classical Coulomb friction model.  
Let $\mathbf{f}_{t,i}$ be the tangential contact force, $\mathbf{n}_i$ the contact normal, and $\mu_i$ the friction coefficient.
Rather than the true friction cone, we use a polyhedral approximation where Coulomb's law constrains the total contact force to lie within a friction pyramid~\cite{anitescu1997formulating}.

\begin{equation}
\scalebox{0.82}{$
\mathcal{F}_C(\boldsymbol{l}) =
\big\{\,
\mathbf{f}_{c,i} = f_{n,i}\mathbf{n}_i + \mathbf{D}_i\boldsymbol{\beta}_i
\;\big|\;
f_{n,i} \ge 0,\;
\boldsymbol{\beta}_i \ge 0,\;
\mathbf{e}^\top\boldsymbol{\beta}_i \le \mu_i f_{n,i}
\big\},$}
\label{eq:friction_set}
\end{equation}
where $\mathbf{D}_i$ spans $r$ tangential directions forming the polyhedral approximation, and $\boldsymbol{\beta}_i \in \mathbb{R}^r_{\ge 0}$ are the nonnegative magnitudes along these directions.  
The inequality $\mathbf{e}^\top\boldsymbol{\beta}_i \le \mu_i f_{n,i}$ defines the boundary of the friction pyramid, separating sticking ($<$) from sliding ($=$) conditions.

The second complementarity condition relates tangential velocity to friction forces. Let $\boldsymbol{\nu}_{t,i}$ denote the relative tangential velocity at contact $i$. The tangential velocity components along the friction directions are complementary to the friction force magnitudes:
\begin{equation}
0 \le \boldsymbol{\nu}_{t,i} + \gamma_i \mathbf{e} \perp \boldsymbol{\beta}_i \ge \mathbf{0},
\label{eq:tang_comp}
\end{equation}
where $\gamma_i \ge 0$ is a slack variable and the complementarity is understood component-wise. When sticking ($\boldsymbol{\nu}_{t,i} = \mathbf{0}$), friction forces $\boldsymbol{\beta}_i$ can be nonzero; when sliding ($\boldsymbol{\nu}_{t,i} \neq \mathbf{0}$), friction is at maximum. 

The third complementarity condition enforces the friction pyramid boundary constraint:
\begin{equation}
0 \le \mu_i f_{n,i} - \mathbf{e}^\top\boldsymbol{\beta}_i \perp \gamma_i \ge 0.
\label{eq:slack_comp}
\end{equation}
When $\gamma_i = 0$, the contact lies strictly inside the friction pyramid (sticking); when $\gamma_i > 0$, the inequality $\mathbf{e}^\top\boldsymbol{\beta}_i = \mu_i f_{n,i}$ becomes active and the contact slides.
Together, these three complementarity conditions enable a unified algebraic treatment of all contact modes (separation, sticking, and sliding) under a single formulation.
  
\vspace{-3 mm}
\subsection{Robot Dynamics with Contact and Global LCP Assembly}
\label{subsec:dyn_contact}

At each contact candidate $i$, we define a \emph{contact frame} as a local coordinate frame with its $z$-axis aligned with the contact normal $\mathbf{n}_i$ and its $x$–$y$ axes spanning the tangent plane. The relative velocity at contact $i$ is $\boldsymbol{\nu}_i = \mathbf{J}_i^{\mathrm{w}} \mathbf{v}$ in world frame, where $\mathbf{J}_i^{\mathrm{w}}$ is the world-frame Jacobian (defined below) and $\mathbf{v} := \dot{\boldsymbol{l}}$ denotes the actuator velocities (notation introduced for brevity in the discrete and contact formulations). The normal component is $\nu_{n,i} = \mathbf{n}_i^\top \boldsymbol{\nu}_i$ and the tangential component $\boldsymbol{\nu}_{t,i}$ is obtained by projecting $\boldsymbol{\nu}_i$ onto the tangent plane.

We write the actuator–space dynamics in discrete formulation with a time step $h$ as follows 
\begin{equation}
\mathbf{M}\big(\mathbf{v}^{\,k+1}-\mathbf{v}^{\,k}\big)
\;=\;
h\,\mathbf{f}_{\text{free}}
\;+\;
\sum_{i=1}^{m}\big(\mathbf{J}_i^{\mathrm{w}}\big)^\top\,\mathbf{f}_{c,i},
\label{eq:disc_dyn_soft}
\vspace{-2 mm}
\end{equation}
where $\mathbf{v}$ are actuator velocities, $m$ is the number of contact candidates,
$\mathbf{J}_i^{\mathrm{w}}\in\mathbb{R}^{3\times n}$ is the world-frame contact Jacobian for candidate $i$, and
$\mathbf{f}_{c,i}\in\mathbb{R}^3$ is the corresponding contact impulse (force $\times$ $h$) in world frame.
The term $\mathbf{f}_{\text{free}}$ collects \emph{all} noncontact contributions from the EoM
(actuation, stiffness, damping, gravity), evaluated at time $k$.
Note that in the discrete-time LCP formulation that follows, the variables $\bar{f}_{n,i}$ and $\bar{\boldsymbol{\beta}}_i$ represent impulses (force $\times$ $h$) rather than instantaneous forces, consistent with the time-stepping formulation. We use barred notation to distinguish impulses from the force variables used in the complementarity conditions above.
We update configuration by $\boldsymbol{l}^{k+1}=\boldsymbol{l}^{k}+h\,\mathbf{v}^{\,k+1}$.

\subsubsection{Contact frame and force decomposition}
At contact $i$, let $\mathbf{R}_i$ be a rotation that aligns the local frame
with normal $\mathbf{n}_i$ (its $z$-axis) and spans the tangent plane ($x$–$y$).
The world-frame Jacobian $\mathbf{J}_i^{\mathrm{w}}\in\mathbb{R}^{3\times n}$ maps actuator velocities to world-frame relative velocities. For computational convenience in the LCP, we also define the contact–frame Jacobian $\mathbf{J}_i = \mathbf{R}_i^\top \mathbf{J}_i^{\mathrm{w}}$, which rotates velocities to the contact frame.
Partition  
\[
\mathbf{J}_i \;=\;
\begin{bmatrix}
\mathbf{J}_{t,i}\\ \mathbf{J}_{n,i}
\end{bmatrix},
\quad
\mathbf{J}_{t,i}\in\mathbb{R}^{2\times n},\ \ \mathbf{J}_{n,i}\in\mathbb{R}^{1\times n}.
\]
Approximate Coulomb friction by $r$ facet directions collected in
$\mathbf{D}_i\in\mathbb{R}^{2\times r}$, and write the contact impulse (force $\times$ $h$) in world frame as
\begin{equation}
\mathbf{f}_{c,i}
\;=\;
\mathbf{R}_i
\begin{bmatrix}
\mathbf{D}_i \bar{\boldsymbol{\beta}}_i\\[2pt] \bar{f}_{n,i}
\end{bmatrix},
\qquad
\bar{\boldsymbol{\beta}}_i\in\mathbb{R}^{r}_{\ge 0},\ \ \bar{f}_{n,i}\ge 0,
\label{eq:fc_decomp}
\end{equation}
where $\bar{f}_{n,i}$ and $\bar{\boldsymbol{\beta}}_i$ represent impulses rather than instantaneous forces, and $\mathbf{R}_i$ rotates from contact frame to world frame. The tangential components are $\mathbf{D}_i \bar{\boldsymbol{\beta}}_i$ and the normal component is $\bar{f}_{n,i}$, both in contact frame before rotation.

\subsubsection{LCP assembly in contact frame.}
For the LCP formulation, we work entirely in contact frame. The contact-frame Jacobian $\mathbf{J}_i = \mathbf{R}_i^\top \mathbf{J}_i^{\mathrm{w}}$ partitions into normal and tangential components $\mathbf{J}_{n,i}$ and $\mathbf{J}_{t,i}$ as defined above. From \eqref{eq:disc_dyn_soft}, projecting the free velocity to contact frame gives
\[
\mathbf{b}^{\text{free}}_i
\;=\;
\mathbf{J}_i
\Big(\mathbf{v}^{\,k} + h\,\mathbf{M}^{-1}\mathbf{f}_{\text{free}}\Big)
=
\begin{bmatrix}
\big[\mathbf{b}^{\text{free}}_i\big]_{xy}\\[2pt] \big[\mathbf{b}^{\text{free}}_i\big]_z
\end{bmatrix},
\]
with scalar normal part $b_{n,i}:=\big[\mathbf{b}^{\text{free}}_i\big]_z$ and tangential part
$\mathbf{b}_{t,i}:=\mathbf{D}_i^\top\big[\mathbf{b}^{\text{free}}_i\big]_{xy}$.

\subsubsection{Per–contact linear maps (self terms).}
Because $\mathbf{v}^{\,k+1}$ depends linearly on all contact impulses,
the $i$-th contact relative velocities are linear in the impulse variables \((\bar{f}_{n,i},\bar{\boldsymbol{\beta}}_i)\).
Define the (self) blocks 
\begin{equation}
\centering
\scalebox{0.8}{$
\begin{alignedat}{2}
A_{nn}^{(i)} &= \mathbf{J}_{n,i}\,\mathbf{M}^{-1}\,\mathbf{J}_{n,i}^\top \in \mathbb{R},
\qquad &
A_{nt}^{(i)} &= \mathbf{J}_{n,i}\,\mathbf{M}^{-1}\,\mathbf{J}_{t,i}^\top\,\mathbf{D}_i \in \mathbb{R}^{1\times r},\\[4pt]
A_{tn}^{(i)} &= \mathbf{D}_i^\top\,\mathbf{J}_{t,i}\,\mathbf{M}^{-1}\,\mathbf{J}_{n,i}^\top \in \mathbb{R}^{r\times 1},
\qquad &
A_{tt}^{(i)} &= \mathbf{D}_i^\top\,\mathbf{J}_{t,i}\,\mathbf{M}^{-1}\,\mathbf{J}_{t,i}^\top\,\mathbf{D}_i \in \mathbb{R}^{r\times r}.
\end{alignedat}
$}
\label{eq:self_blocks}
\end{equation}

With the cone–boundary slack $\gamma_i\ge 0$, collect unknowns and slacks as
\[
\mathbf{z}_i := \begin{bmatrix} \bar{f}_{n,i}\\ \bar{\boldsymbol{\beta}}_i\\ \gamma_i \end{bmatrix}
\in\mathbb{R}^{1+r+1}_{\ge 0},
\qquad
\mathbf{w}_i := \begin{bmatrix} \nu_{n,i}\\ \mathbf{w}_{t,i}\\ w_{g,i} \end{bmatrix}
\in\mathbb{R}^{1+r+1}_{\ge 0}.
\]
Let $\mathbf{e} = [1,1,\ldots,1]^\top \in \mathbb{R}^r$ be the vector of ones. The LCP variables $\mathbf{w}_i$ encode the three complementarity conditions from Section~\ref{subsec:CompForm}: $\mathbf{w}_{t,i}$ corresponds to $\boldsymbol{\nu}_{t,i} + \gamma_i \mathbf{e}$ in \eqref{eq:tang_comp}, and $w_{g,i}$ corresponds to $\mu_i \bar{f}_{n,i} - \mathbf{e}^\top\bar{\boldsymbol{\beta}}_i$ in \eqref{eq:slack_comp}.
The per–contact (self) relation that encodes dynamics \emph{and} the friction‐cone
complementarities is
\begin{equation}
\underbrace{\begin{bmatrix}
\nu_{n,i}\\ \mathbf{w}_{t,i}\\ w_{g,i}
\end{bmatrix}}_{\mathbf{w}_i}
\;=\;
\underbrace{\begin{bmatrix}
A_{nn}^{(i)} & A_{nt}^{(i)} & \mathbf{0} \\[3pt]
A_{tn}^{(i)} & A_{tt}^{(i)} & \mathbf{e} \\[3pt]
\mu_i & -\mathbf{e}^\top & 0
\end{bmatrix}}_{\mathbf{A}_i}
\underbrace{\begin{bmatrix}
\bar{f}_{n,i}\\ \bar{\boldsymbol{\beta}}_i\\ \gamma_i
\end{bmatrix}}_{\mathbf{z}_i}
\;+\;
\underbrace{\begin{bmatrix}
b_{n,i}\\ \mathbf{b}_{t,i}\\ 0
\end{bmatrix}}_{\mathbf{b}_i},
\label{eq:self_block}
\end{equation}
where the \((2,3)\) block \(\mathbf{e}\) and the bottom row \([\mu_i\;-\mathbf{e}^\top\;0]\) are the
standard linearization of the Coulomb cone and its complementarity with \(\gamma_i\).
% \yc{what is this}

\paragraph*{Cross–contact coupling (off–diagonal blocks).}
Velocities at contact $i$ also depend on reactions at $j\neq i$ through the same linear map,
yielding off–diagonal blocks
\begin{equation}
\mathbf{A}_{ij}
=
\begin{bmatrix}
A_{nn}^{(ij)} & A_{nt}^{(ij)} & \mathbf{0}\\[3pt]
A_{tn}^{(ij)} & A_{tt}^{(ij)} & \mathbf{0}\\[3pt]
\mathbf{0} & \mathbf{0} & 0
\end{bmatrix},
\qquad
\begin{aligned}
A_{nn}^{(ij)} &= \mathbf{J}_{n,i}\mathbf{M}^{-1}\mathbf{J}_{n,j}^\top,\\
A_{nt}^{(ij)} &= \mathbf{J}_{n,i}\mathbf{M}^{-1}\mathbf{J}_{t,j}^\top\mathbf{D}_j,\\
A_{tn}^{(ij)} &= \mathbf{D}_i^\top\mathbf{J}_{t,i}\mathbf{M}^{-1}\mathbf{J}_{n,j}^\top,\\
A_{tt}^{(ij)} &= \mathbf{D}_i^\top\mathbf{J}_{t,i}\mathbf{M}^{-1}\mathbf{J}_{t,j}^\top\mathbf{D}_j.
\end{aligned}
\label{eq:cross_blocks}
\end{equation}
Intuitively: a normal/tangential impulse at $j$ changes $\mathbf{v}^{\,k+1}$,
which changes both $\nu_{n,i}$ and $\boldsymbol{\nu}_{t,i}$ through $\mathbf{J}_{n,i},\mathbf{J}_{t,i}$.

\paragraph*{Global LCP (final form).}
Stack all $m$ contact candidates as
\[
\mathbf{z} := \begin{bmatrix}\mathbf{z}_1\\ \vdots\\ \mathbf{z}_m\end{bmatrix},
\quad
\mathbf{w} := \begin{bmatrix}\mathbf{w}_1\\ \vdots\\ \mathbf{w}_m\end{bmatrix},
\quad
\mathbf{b} := \begin{bmatrix}\mathbf{b}_1\\ \vdots\\ \mathbf{b}_m\end{bmatrix}.
\]
The complete linear complementarity system is
\begin{equation}
\mathbf{w}
=
\underbrace{\begin{bmatrix}
\mathbf{A}_1 & \cdots & \mathbf{A}_{1m}\\
\vdots & \ddots & \vdots\\
\mathbf{A}_{m1} & \cdots & \mathbf{A}_m
\end{bmatrix}}_{\mathbf{A}\in\mathbb{R}^{m(1+r+1)\times m(1+r+1)}}
\mathbf{z}
+\mathbf{b},
\qquad
\mathbf{w}\ge \mathbf{0},\ \mathbf{z}\ge \mathbf{0},\ \mathbf{w}^\top\mathbf{z}=0.
\label{eq:lcp_global}
\end{equation}
% Diagonal blocks \(\mathbf{A}_i\) are the self terms \eqref{eq:self_block};
% off–diagonal blocks \(\mathbf{A}_{ij}\) are the cross terms \eqref{eq:cross_blocks}.
% The ones/minus–ones pattern in \(\mathbf{A}_i\) is precisely the friction–cone linearization
% (\(\mathbf{E}\), \(\mathbf{e}^\top\)) and its complementarity with \(\gamma_i\).
% All inertial/geometric couplings (self and cross) enter via \(\mathbf{M}^{-1}\) and the Jacobians.

% \paragraph*{Practical note.}
% In implementation, we compute \(\mathbf{A}\) from \(\mathbf{M}^{-1}\) using a Cholesky factor
% (\(\mathbf{M}=\mathbf{L}\mathbf{L}^\top\)) for stability, assemble \(\mathbf{b}\) from the
% contact–frame free velocities \((b_{n,i},\mathbf{b}_{t,i})\), and then solve
% \eqref{eq:lcp_global} with a standard LCP backend (e.g., PATH or Lemke).
\vspace{-5 mm}
\subsection{LCP Reformulation and Numerical Conditioning}
\label{sec:LCP_Reformulation}

The global LCP assembly derived in \eqref{eq:lcp_global} provides a physically consistent description of multi-contact dynamics. However, the direct numerical solution of this system in soft robotics is prone to failure. This is not due to solver implementation, but rather a structural characteristic of discretized continuum bodies: \textit{Continuum Redundancy}.

When a soft robot section interacts with the environment, many collision geometries (disks) along the body can be simultaneously active within a section or across the whole robot. This can create a statically indeterminate system where the contact Jacobian rows become effectively linearly dependent relative to the available generalized coordinates, rendering the LCP matrix $\mathbf{A}$ singular. Furthermore, the disparity between high pneumatic stiffness terms 
and low friction coefficients 
introduces severe ill-conditioning.

To resolve these structural pathologies without sacrificing physical accuracy, we propose a three-stage conditioning pipeline applied \emph{before} calling an LCP backend: (1) inertial normal-row rank selection, (2) Ruiz equilibration of the coupled system, and (3) light Tikhonov regularization on the remaining normal blocks.
This makes the formulation robust so that different off-the-shelf LCP algorithms can be used effectively.

\subsubsection{Stage 1: Inertial Rank Selection (Whitened RRQR)}
Standard rank-reduction techniques often rely on purely geometric heuristics (e.g., removing contacts with similar outward normals). In dynamic simulation, however, geometry alone is insufficient: contacts at the base and at the tip may share a normal direction but have very different effective inertias.
To identify \emph{dynamically} redundant contacts, we first stack all candidate normal Jacobians into $\mathbf{J}_n \in \mathbb{R}^{m\times n}$ and ``whiten'' them in the inertia-weighted metric induced by the robot inertia $\mathbf{M}=\mathbf{L}\mathbf{L}^\top$:
\begin{equation}
    \mathbf{S} = \mathbf{J}_n \mathbf{L}^{-\top} \in \mathbb{R}^{m\times n}.
\end{equation}
Here, $\mathbf{L}$ is the Cholesky factor of the inertia matrix $\mathbf{M}$ (see \eqref{eq:EoM}); in these whitened coordinates the inertia becomes the identity, so $\mathbf{S}$ normalizes each contact row by its inertial ``leverage'' and near-redundant constraints become nearly linearly dependent in this space.
We then perform Rank-Revealing QR (RRQR) with column pivoting on the transpose of the whitened matrix:
\begin{equation}
    \mathbf{S}^\top \mathbf{P} = \mathbf{Q} \mathbf{R},
\end{equation}
where $\mathbf{P}$ is a permutation matrix. Let $\eta = \varepsilon_{\text{rank}} |R_{11}|$ be a tolerance; all diagonals $|R_{ii}| \le \eta$ are treated as dynamically redundant. The corresponding rows in $\mathbf{J}_n$ are discarded, defining a ``keep set'' of normal directions used to assemble the coupled friction LCP. Whenever a normal row is removed, we simultaneously discard its associated friction directions so that the retained normal rows are linearly independent, removing the structural singularity from redundant contacts while keeping the reduced LCP physically consistent.

\subsubsection{Stage 2: Ruiz Equilibration}
Even after rank selection, the resulting LCP matrix mixes quantities with very different units and magnitudes (accelerations, forces, and friction slacks), which leads to poor conditioning.
To mitigate this, we apply iterative Ruiz equilibration~\cite{ruiz2001scaling} to $(\mathbf{A},\mathbf{b})$: at each iteration $\kappa$, we rescale rows and columns using diagonal matrices $\mathbf{S}_r^{(\kappa)}, \mathbf{S}_c^{(\kappa)}$ based on their $\ell_2$ norms, and update
\[
 \mathbf{A}^{(\kappa+1)} = \mathbf{S}_r^{(\kappa)} \mathbf{A}^{(\kappa)} \mathbf{S}_c^{(\kappa)}, 
 \qquad
 \mathbf{b}^{(\kappa+1)} = \mathbf{S}_r^{(\kappa)} \mathbf{b}^{(\kappa)}.
\]
After a small fixed number of iterations, we obtain a scaled system
\(
 \mathbf{A}_{\text{scaled}} = \mathbf{S}_r \mathbf{A} \mathbf{S}_c,\ 
 \mathbf{b}_{\text{scaled}} = \mathbf{S}_r \mathbf{b}
\)
whose rows and columns have comparable norms. This balances the dynamic range across normal and tangential blocks and significantly improves pivoting behavior in the downstream LCP solver.

\subsubsection{Stage 3: Tikhonov Regularization}
Finally, to guarantee strictly positive pivots on the normal-force variables, we add a light Tikhonov regularization only on the corresponding diagonal entries of the scaled matrix:
\begin{equation}
    \mathbf{A}_{\text{final}} = \mathbf{A}_{\text{scaled}} + \varepsilon_W \mathbf{I}_n^{(N)},
\end{equation}
where $\mathbf{I}_n^{(N)}$ is the identity restricted to the $N$ scalar normal-force variables.
Because rank deficiency has already been removed by Stage~1 and scale disparities are handled by Stage~2, this $\varepsilon_W$ can be chosen extremely small: it does not alter contact forces in practice, but it guarantees positive definiteness of each normal block and robust convergence of the LCP backend.

\vspace{-3 mm}
\section{Planning Through Contact} 
\label{sec:Planning}

Building on the forward simulation framework of Section~\ref{sec:Forward}, we now address the inverse problem: planning trajectories for soft robots through contact-rich manipulation tasks.
While the simulation framework supports multi-contact interactions along the entire continuum body, trajectory optimization through contact introduces additional computational challenges due to the nonconvexity and combinatorial nature of contact mode selection.
As a first step toward contact-aware planning for soft robots, we focus here on manipulation tasks where contact occurs primarily at the end-effector disk. 
This restriction reduces problem dimensionality while still capturing the essential challenges of contact mode selection, friction constraints, and intermittent contact transitions.
The formulation naturally extends to multi-contact scenarios, but we defer a full treatment of whole-body contact planning to future work.

A key contribution of this work is a unified model that supports both the direct problem (forward simulation) and the inverse problem (trajectory optimization) without changing the contact mechanics.
We pose planning as a trajectory optimization problem using direct transcription, treating the trajectory of states, inputs, and contact forces as decision variables and encoding contact via complementarity constraints.
This yields an MPCC \cite{posa2014direct, underactuated} that remains consistent with the simulation model and naturally supports intermittent contacts. 
\vspace{-7 mm}
\subsection{Problem Formulation and Dynamics}
\textbf{Robot dynamics.}
We employ a single-step implicit (backward) Euler discretization to improve stability and enable larger time steps than explicit schemes.
Building on \eqref{eq:disc_dyn_soft}, the implicit discrete dynamics (in actuator space) are
\begin{equation}
\begin{aligned}
\boldsymbol{l}_{k+1} &= \boldsymbol{l}_k + h\,\boldsymbol{v}_{k+1}, \\
\mathbf{M}_{k+1}\big(\boldsymbol{v}_{k+1} - \boldsymbol{v}_k\big) &= h \Big(\boldsymbol{f}^{\,\text{free}}_{k+1} + \boldsymbol{J}_{k+1}^{\top}\,\boldsymbol{f}_{k+1}\Big),
\end{aligned}
\label{eq:mpcc-dynamics}
\end{equation}
where \(\mathbf{M}\) is the inertia matrix and \(\boldsymbol{J}\) the contact Jacobian.
Consistent with \eqref{eq:EoM}, the free contribution groups all noncontact terms evaluated at \(k{+}1\):
\[
\boldsymbol{f}^{\,\text{free}}_{k+1} := \boldsymbol{\tau}_{k+1} - \boldsymbol{C}_{k+1}\boldsymbol{v}_{k+1} - \boldsymbol{K}\,\boldsymbol{l}_{k+1} - \boldsymbol{B}\,\boldsymbol{v}_{k+1} - \boldsymbol{G}_{k+1}.
\]
\begin{figure*}[t]
  \centering
  \includegraphics[width=\linewidth]{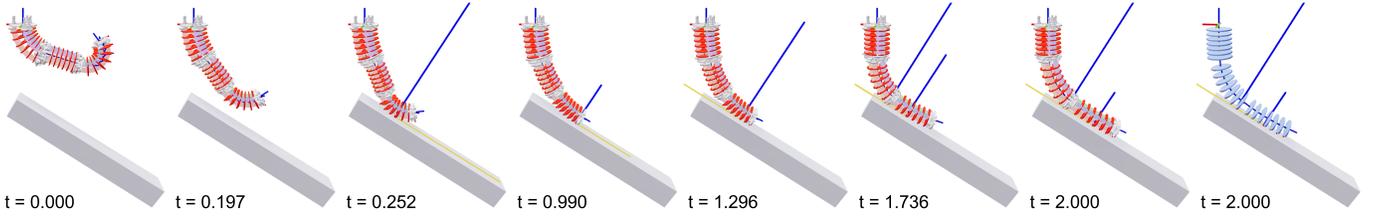}
  \caption{Forward simulation of the soft robot onto an inclined rigid box ($t \in [0, 2]$~s) with six intermediate disks per section as contact candidates, demonstrating that the framework supports an arbitrary number of disks. The robot free-falls from $t = 0$--$1$~s, then actuation ramps on sections~2 and~3 from $t = 1$--$2$~s.}
  \label{fig:NewSliding6Disk}
  \vspace{-5 mm}
\end{figure*}
\textbf{Ball dynamics.} 
The manipulated object is a homogeneous sphere with inertia $\boldsymbol{I}_{\text{ball}}$. 
Its discretized rotational dynamics are
\begin{equation}
    \boldsymbol{I}_{\text{ball}} (\boldsymbol{\omega}_{\text{ball},k+1} - \boldsymbol{\omega}_{\text{ball},k}) = h \left(\boldsymbol{J}_{b,k+1}^\top \boldsymbol{f}_{k+1} - \boldsymbol{D}_{\text{ball}} \boldsymbol{\omega}_{\text{ball},k+1}\right),
    \label{eq:ball-dynamics}
\end{equation}
where $\boldsymbol{J}_{b}$ maps contact forces $\boldsymbol{f}_{k+1}$ to object torques.
Orientation updates via $\boldsymbol{q}_{\text{ball},k+1} = \exp(h \boldsymbol{\omega}_{\text{ball},k+1}) \odot \boldsymbol{q}_{\text{ball},k}$, where $\odot$ denotes quaternion multiplication.
The quaternion exponential map implementation uses a smooth approximation to avoid numerical singularities, as detailed in Subsection~\ref{subsec:proposed_method}.

\textbf{Contact complementarity constraints.}
Contact is modeled using complementarity constraints consistent with Section~\ref{subsec:CompForm}, adapted here for trajectory optimization.
Unlike the velocity-level complementarity used in forward simulation (Section~\ref{sec:Forward}), we employ a gap--force formulation to improve numerical conditioning in the optimization.
For each time interval $k\in\{0,\ldots,N-1\}$, let $f_{k,z}\ge 0$ denote the normal contact force, 
$\boldsymbol{f}_{k,t}\in\mathbb{R}^d_{\ge 0}$ the tangential force magnitudes along the $d$ friction pyramid directions collected in $\boldsymbol{D}$,
and $\gamma_k\ge 0$ the friction slack variable.
The signed gap function $\phi(\boldsymbol{l}_{k+1})$ and the relative tangential contact velocity $\boldsymbol{\nu}_{t,k+1}$ are evaluated at the end of the interval using the configuration $\boldsymbol{l}_{k+1}$ and velocities $\boldsymbol{v}_{k+1}$, with tangential components projected onto the friction directions via $\boldsymbol{D}^\top \boldsymbol{\nu}_{t,k+1}$. Although $\phi(\boldsymbol{l}_{k+1})$ depends on velocity through integration, avoiding direct force–velocity coupling improves numerical conditioning.
The resulting non-penetration, friction cone, and stick--slip conditions are enforced through the following complementarity system:

\begin{equation}
\setlength{\jot}{1pt}
\begin{aligned}
&0 \le \phi(\boldsymbol{l}_{k+1}) \perp f_{k,z} \ge 0,\\
&0 \le \mu f_{k,z} - \mathbf{e}^\top \boldsymbol{f}_{k,t} \perp \gamma_k \ge 0,\\
&\boldsymbol{0} \le \boldsymbol{f}_{k,t} \perp 
\boldsymbol{D}^\top \boldsymbol{\nu}_{t,k+1} + \gamma_k \mathbf{e} \ge \boldsymbol{0},
\end{aligned}
\label{eq:ComConst}
\end{equation}
These constraints encode the same contact modes (separation, sticking, and sliding) as in Section~\ref{subsec:CompForm}, while using a gap--force formulation that is better conditioned for trajectory optimization.

\vspace{-3 mm}
\subsection{Proposed Solution Method}
\label{subsec:proposed_method}

Solving the MPCC for a 9-DoF soft robot with highly nonlinear dynamics is challenging: complementarity introduces nonconvexity and a combinatorial contact-mode space at each time step, and the number of decision variables (states, controls, contact forces over the horizon) is large, so cold-start optimization often fails or converges poorly.
Our approach is a two-stage warm-start that uses kinematic planning to initialize the full dynamic optimization and thereby guides the optimizer to a feasible, contact-consistent region.

In the first stage we solve a kinematic contact problem in which robot dynamics are ignored while ball dynamics and all contact constraints are preserved.
This problem is easier to solve because it removes the robot's inertial coupling and shrinks the search space, yet it still captures contact interactions and manipulation objectives; the solution yields a feasible trajectory and the contact mode sequence (separation, sticking, sliding over time).
In the second stage we warm-start the full dynamic optimization from that solution: we initialize states \((\boldsymbol{l}_k, \boldsymbol{v}_k, \boldsymbol{q}_{\text{ball},k}, \boldsymbol{\omega}_{\text{ball},k})\) and contact variables \((f_{k,z}, f^i_{k,t}, \gamma_k)\) from the kinematic result, and we obtain control torques \(\boldsymbol{\tau}_k\) by inverting the robot dynamics along that trajectory so the initial guess is dynamically consistent.
This warm-start dramatically improves convergence and solution quality compared to cold starts.

To enable this strategy we address several implementation details.

(i) Contact detection between the end-effector disk and the ball uses a signed distance field (SDF): the disk is a capped cylinder and the ball a sphere. The gap \(\phi(\boldsymbol{l}_k)\) is built from this SDF; because \(\max(\cdot,\cdot)\) and \(|\cdot|\) are non-differentiable and would break gradient-based optimization, we replace them with smooth surrogates: \(\text{sabs}(s) = \sqrt{s^2 + \varepsilon_{\text{sdf}}}\) for the absolute value and \(\text{smax}(s_1,s_2) = \tfrac{1}{2}(s_1 + s_2 + \sqrt{(s_1-s_2)^2 + \varepsilon_{\text{sdf}}})\) for the maximum, with \(\varepsilon_{\text{sdf}} \approx 10^{-12}\). This yields a differentiable gap and contact normals \(\boldsymbol{n}_k = \nabla_{\boldsymbol{l}_k} \phi(\boldsymbol{l}_k)\). The full SDF expression and gap definition are given in the released code.

(ii) The ball's quaternion update uses $\exp(h\boldsymbol{\omega})$, which divides by $\theta = \|h\boldsymbol{\omega}\|$ and is singular at $\boldsymbol{\omega} \to 0$, so gradient computation fails there.
We use a smooth Taylor-series approximation \cite{sola2017quaternion} so the exponential remains differentiable and unit norm is preserved.

(iii) We solve the MPCC with IPOPT~\cite{wachter2006implementation} and Scholtes relaxation~\cite{scholtes2001convergence}: each complementarity product in \eqref{eq:ComConst} is replaced by \(a \cdot b \le \varepsilon\) with \(\varepsilon \approx 10^{-7}\), yielding a standard NLP that approximately enforces complementarity.

\vspace{-2 mm}
\section{Results}
\label{sec:Results}

This section reports results in three parts. We first describe the physics-engine pipeline for robot modeling and contact resolution using the proposed formulation. We then evaluate forward simulation performance. Finally, we present planning through contact, illustrated by a ball manipulation task, highlighting the applicability and generalizability of the approach for both forward simulation and trajectory optimization through contact.

\vspace{-2 mm}
\subsection{Physics Engine}
All entities are expressed in a common world frame with consistent units, and robots and objects expose kinematics and dynamics through a unified API (positions, velocities, Jacobians, and inertial properties). Dynamics and Jacobians are generated using CasADi and compiled to C for fast, differentiable evaluation shared across simulation and planning \cite{andersson2019casadi}. Geometry is represented using convex primitives, with robot contact elements (e.g., soft-robot disks) and environment obstacles registered as convex shapes; contact detection and signed-distance queries are performed using Minkowski Portal Refinement (MPR) \cite{ericson2004real,eberly20063d,Fabisch_distance3d_Distance_computation}. At each time step, the simulator advances dynamics using semi-implicit Euler with $h=10^{-4}$~s (Scenario~C additionally evaluates RK23 with $h=10^{-3}$~s), queries candidate contact pairs and signed gaps, constructs contact frames and Jacobians, assembles and conditions the per-step linear complementarity problem (LCP) as in Section~\ref{sec:Forward}, and solves it using a semismooth Newton method based on the Fischer--Burmeister reformulation with a Levenberg--Marquardt trust region \cite{fischer1995newton,bazaraa2006nonlinear}. The resulting contact forces are mapped back to generalized coordinates, with optional gap stabilization adding $(\alpha/h)\,g_i$ to the normal right-hand side to reduce penetration drift, and visualization is handled using Viser \cite{yi2025viser}.

\vspace{-3 mm}
\subsection{Forward Simulation}
\label{sec:res_forward}

We evaluate forward simulation using the proposed physics engine and contact pipeline on a three-section pneumatically actuated soft arm (PCC model of Section~\ref{sec:modeling}) with $L_0 = 0.15$~m, section masses $[1.17, 0.54, 0.265]$~kg, stiffness $K = 265$~N/m, and damping $D = 125$~Ns/m. Candidate contact frames are registered along the backbone, with frames at the base, tip, and multiple interior locations per section; both coarse (3 interior frames) and dense (6 interior frames) discretizations are evaluated to induce varying degrees of whole-body contact.

\noindent\textbf{Scenario~A: Inclined rigid box.}
We consider a challenging whole-body contact scenario using six intermediate disks per section, yielding a dense contact discretization along the backbone. The robot free-falls onto an inclined rigid box, where elastic relaxation of a pre-elongated configuration, governed by the section stiffness, induces a brief axial contraction and alignment with the surface without penetration. At $t=1$~s, a linear torque ramp (up to $40$~N$\cdot$m) is applied to sections~2 and~3, while section~1 remains unactuated. As actuation increases, the robot transitions from sticking to sliding, demonstrating non-penetrating frictional contact under varying surface normals (Fig.~\ref{fig:NewSliding6Disk}).

\begin{figure}[t]
  \centering
  \includegraphics[width=\linewidth]{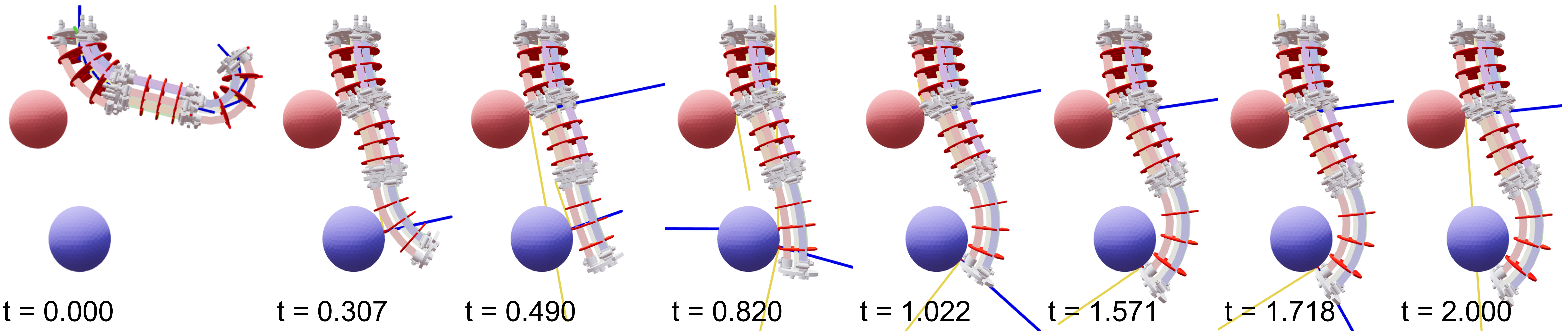}
\caption{Scenario~B: sequential contact and whole-body deformation of a soft robot interacting with two spherical obstacles under distal actuation.}
  \label{fig:TwoBall}
  \vspace{-7 mm}
\end{figure}

\noindent\textbf{Scenario~B: Two spherical obstacles.}
This scenario replaces the inclined box in Scenario~A with two spherical obstacles, while using three disks per section. The robot first establishes contact with the upper sphere, after which torque actuation at the distal joints (joints~7 and~8) induces bending toward a second, lower sphere. As contact is formed with the lower obstacle, the resulting contact forces propagate along the backbone, producing whole-body deformation captured by the proposed pipeline (Fig.~\ref{fig:TwoBall}).

\noindent\textbf{Scenario~C: Ablation study.}
We conduct an ablation study to assess the contribution of each stage of the proposed conditioning pipeline and to justify its design choices when using the semismooth Newton LCP solver. The setup mirrors Scenario~A, and we run 64 experiments spanning two box angles (45° and 60°), two contact discretizations (3 and 6 disks per section), and two integrators (Euler with $h=10^{-4}$~s and RK23 with $h=10^{-3}$~s). We compare a baseline configuration without conditioning against individual components (inertial rank-revealing QR, Ruiz equilibration, and Tikhonov regularization), all pairwise combinations, and the full pipeline. Success is defined by an LCP convergence rate of at least 95\%, maximum penetration below 10~mm, and simulation completion of at least 80\%. Without Ruiz equilibration, the success rate drops to 41\% (12/29 experiments), with frequent solver timeouts and incomplete simulations. In contrast, all configurations that include Ruiz equilibration succeed, and the full three-stage pipeline improves the success rate from 43\% (baseline) to 100\%, achieving reliable convergence and completion across all tested conditions. These results indicate that Ruiz equilibration is essential for robustness, while the complete pipeline (QR, Ruiz, Tikhonov) provides the most reliable overall performance.

\vspace{-3 mm}
\subsection{Trajectory Optimization Through Contact}
\label{sec:res_planning}

We evaluate the proposed MPCC formulation on contact-rich ball manipulation tasks using the 9-DoF soft robot. The task consists of rotating a fixed-position sphere (radius $0.05$~m, centered at $[0.0,\,0.1,\,0.7]^\top$~m) by $90^\circ$ about a specified axis while maintaining continuous contact through the end-effector disk. Three scenarios are considered, corresponding to rotations about the $x$, $y$, and $z$ axes. The optimization uses $N=100$ time steps over a $T=4.0$~s horizon, with friction coefficient $\mu=0.6$, $d=10$ friction directions, and Scholtes relaxation parameter $\varepsilon=10^{-7}$.

\paragraph*{Cost function.}
The objective balances task completion, contact consistency, and control effort. Running and terminal quaternion costs penalize orientation error of the ball, while a small reward on the normal contact force encourages sustained contact. Additional terms promote smooth contact forces and velocities, regularize control effort, and softly enforce unit-norm constraints on the ball quaternion. The terminal orientation cost places stronger emphasis on achieving the desired final rotation.
\vspace{-2 mm}
\begin{equation}
\scalebox{0.85}{$
\begin{aligned}
J = &\sum_{k=0}^{N-1} h \Big[ 10\,J_{\text{quat}}(\boldsymbol{q}_{k+1}, \boldsymbol{q}_{\text{target}})
- 10^{-3} f_{k,z}
+ 0.2 (f_{k,z} - f_{k-1,z})^2 \\
&\quad + 10^{-3} \|\boldsymbol{v}_{k+1} - \boldsymbol{v}_k\|^2
+ 10^{-3} \|\boldsymbol{\omega}_{k+1} - \boldsymbol{\omega}_k\|^2
+ 10^{-2} \|\boldsymbol{\tau}_{k+1}\|^2 \\
&\quad + 10 (\|\boldsymbol{q}_{k+1}\|^2 - 1)^2 \Big]
+ 100\,J_{\text{quat}}(\boldsymbol{q}_{N}, \boldsymbol{q}_{\text{target}}),
\end{aligned}
$}
\label{eq:cost_function}
\end{equation}
where $\boldsymbol{q}_k = \boldsymbol{q}_{\text{ball},k}$, $\boldsymbol{\omega}_k = \boldsymbol{\omega}_{\text{ball},k}$, $J_{\text{quat}}(\boldsymbol{q}, \boldsymbol{q}_{\text{target}}) = 1 - (\boldsymbol{q}^\top \boldsymbol{q}_{\text{target}})^2$ penalizes orientation error.

\paragraph*{Two-stage warm-start strategy.}
Following Section~\ref{sec:Planning}, we adopt a two-stage warm-start strategy. We first solve a kinematic optimization in which the robot is treated as kinematic while the ball retains full rotational dynamics, yielding a feasible contact sequence and ball rotation trajectory. This solution is then used to initialize the full dynamic MPCC, including robot states, ball states, and contact forces, with initial control torques obtained via inverse dynamics.

\begin{figure}[t]
  \centering
  \includegraphics[width=\linewidth]{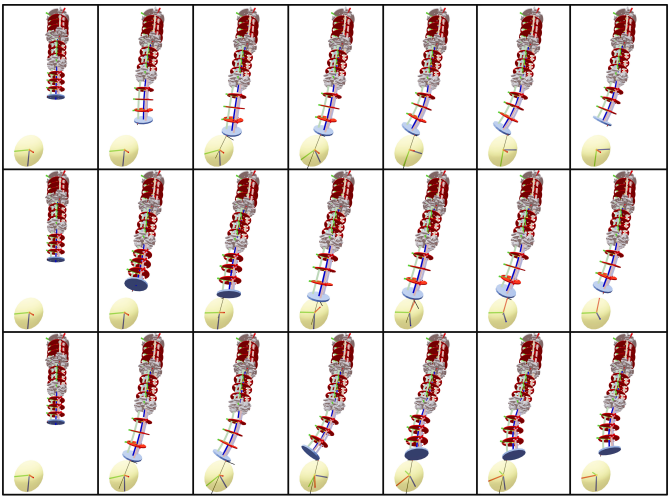}
  \caption{Trajectory optimization results for ball manipulation using the MPCC formulation. Each row corresponds to a $90^\circ$ rotation about a different axis: $x$ (top), $y$ (middle), and $z$ (bottom). Key frames from each optimized trajectory are shown. Coordinate axes: red ($x$), green ($y$), blue ($z$).}
  \label{fig:MPCCResults}
  \vspace{-6 mm}
\end{figure}

\paragraph*{Results.}
Figure~\ref{fig:MPCCResults} shows the optimized trajectories for all three rotation axes. The kinematic stage consistently provides a feasible contact sequence, which the dynamic stage refines into physically consistent motions that respect both inertial dynamics and contact constraints. Across all scenarios, the final quaternion error decreases from $\mathcal{O}(1)$ to approximately $10^{-5}$--$10^{-6}$, indicating near-perfect alignment with the desired $90^\circ$ rotations. 
For the $z$-axis rotation, where the robot lacks a wrist and cannot directly generate twist, the optimizer discovers a non-obvious but physically consistent motion that leverages whole-body dynamics and contact interactions to achieve the task. Average solve times are approximately $4.7$~min for the kinematic stage and $8.9$~min for the dynamic stage (about $13.6$~min total per problem). Detailed quantitative results, including orientation errors, complementarity satisfaction, and solver statistics, are provided in the supplementary material.

\vspace{-3 mm}
\section{Conclusion}
\label{sec:Con}

This work establishes the feasibility of complementarity-based contact modeling and trajectory optimization for soft robotic manipulators.
We have demonstrated that the three-stage conditioning pipeline enables stable simulation of dense multi-contact interactions, and that the kinematically guided warm-start strategy makes dynamic trajectory optimization through contact tractable for soft robots.
The key contribution is proving that such problems are solvable and can produce physically consistent, dynamically feasible trajectories.
This framework provides a foundation for contact-rich manipulation, while adaptive control strategies can address material uncertainties and unmodeled dynamics in practice.
While this paper focuses on establishing feasibility and demonstrating the framework's capabilities, future work will address computational efficiency and real-time control integration.
Specifically, reducing solve times to enable online replanning and integrating the optimization framework with feedback control policies remain important directions for practical deployment.

\ifCLASSOPTIONcaptionsoff
  \newpage
\fi

\bibliographystyle{IEEEtran}
\bibliography{references}

\end{document}

%% file: preamble.tex
\usepackage{algorithm}
\usepackage{algpseudocode}
\usepackage{gensymb}
\usepackage{amsmath}
\usepackage{wrapfig}
\usepackage{hyperref}
\usepackage{graphicx}
\usepackage{epstopdf}
\graphicspath{ {./images/} }
 \usepackage{booktabs} 
\usepackage{adjustbox}
\usepackage{graphics} % for pdf, bitmapped graphics files
\usepackage{epsfig} % for postscript graphics files
\usepackage{times} % assumes new font selection scheme installed
\usepackage{amsmath} % assumes amsmath package installed

\usepackage{amssymb}  % assumes amsmath package installed
\usepackage[utf8]{inputenc}
\usepackage[T1]{fontenc}
\usepackage{mathtools}
\usepackage{enumitem}
\usepackage{float}
\usepackage{gensymb}
\usepackage[font=footnotesize]{caption}
\usepackage{subcaption}

\usepackage{booktabs}

\usepackage{comment}
\usepackage{textcomp}

\newcommand{\cusst}[1]{{}}
\PassOptionsToPackage{table,xcdraw}{xcolor}
\usepackage{xcolor}

\definecolor{Gray}{gray}{0.85}

\usepackage{soul}
\usepackage{cite}

\ifCLASSINFOpdf
\else
\fi